\documentclass[journal]{IEEEtran}

\usepackage{algorithmic}
\usepackage[cmex10]{amsmath}
\usepackage{amsfonts}
\usepackage{blindtext}
\usepackage{cite}
\usepackage{graphicx}
\usepackage{url}
\usepackage{subfigure}

\begin{document}

\title{Event-based Feature Extraction using Adaptive Selection Thresholds}

\author{Saeed~Afshar,
        Ying Xu,
        Jonathan~Tapson,
        Andr\'e~van~Schaik,
        and~Gregory~Cohen,% <-this % stops a space
\thanks{S. Afshar, J. Tapson, A. van~Schaik and G. Cohen are with the MARCS Institute, Western Sydney University, Sydney Australia, 2747.}% <-this % stops a space
\thanks{Email: s.afshar@westernsydney.edu.au.}%
\thanks{Manuscript received February 27, 2019}}

\maketitle

\begin{abstract}
%\boldmath
Unsupervised feature extraction algorithms form one of the most important building blocks in machine learning systems. These algorithms are often adapted to the event-based domain to perform online learning in neuromorphic hardware. However, not designed for the purpose, such algorithms typically require significant simplification during implementation to meet hardware constraints, creating trade offs with performance. Furthermore, conventional feature extraction algorithms are not designed to generate useful intermediary signals which are valuable only in the context of neuromorphic hardware limitations.
In this work a novel event-based feature extraction method is proposed that focuses on these issues. The algorithm operates via simple adaptive selection thresholds which allow a simpler implementation of network homeostasis than previous works by trading off a small amount of information loss in the form of missed events that fall outside the selection thresholds. The behavior of the selection thresholds and the output of the network as a whole are shown to provide uniquely useful signals indicating network weight convergence without the need to access network weights. A novel heuristic method for network size selection is proposed which makes use of noise events and their feature representations. The use of selection thresholds is shown to produce network activation patterns that predict classification accuracy allowing rapid evaluation and optimization of system parameters without the need to run back-end classifiers. The feature extraction method is tested on both the N-MNIST benchmarking dataset and a dataset of airplanes passing through the field of view. Multiple configurations with different classifiers are tested with the results quantifying the resultant performance gains at each processing stage.
\end{abstract}


\IEEEpeerreviewmaketitle

\section{Introduction}
\label{sec:intro}

Feature detection is a fundamental building block required for a wide range of computer vision tasks. These tasks rely on computationally efficient algorithms capable of detecting features in a reliable, repeatable, and robust manner. These features often need to be detected and recognized through a range of transformations, both photometric and affine. Some examples of computer vision tasks which build upon effective feature detection methods include object recognition~\cite{Lowe2004}, image stitching \cite{Snavely2008} and scene classification \cite{Yang2007}. 

The process of feature detection can also be extended to find stable features in video streams~\cite{Laptev2006}, in which the features need to be stable across time as well. These techniques have been used in offline tasks such as video analysis \cite{Dollar2005}, and for more demanding online tasks such as camera localization and visual mapping \cite{Se2002}. 

Neuromorphic vision sensors, also known as silicon retinas \cite{Posch2014}, provide a different imaging paradigm in which to perform computer vision tasks. These devices do not produce the conventional image frames on which conventional computer vision algorithms rely and instead produce data in the form of an asynchronous stream of events with a high temporal resolution. These devices differ on the pixel level from conventional cameras and include specialized hardware circuits at each pixel to generate events only in response to changes in log intensity, providing these devices with a slew of capabilities and operating characteristics that are not possible with conventional imaging devices. 

Conventional feature extraction is an active and well-researched field of study and has produced sophisticated and robust algorithms. Event-based feature extraction poses a different and perhaps more challenging task.
The output event-based cameras constitutes neither a frame, as in traditional computer vision, nor a stream of frames, as in conventional video. As a result, the majority of existing feature detectors are a poor fit for event-based vision data, and requiring the events to be converted into standard image frames before processing. In contrast, this work explores the use of a novel technique to extract useful event-based features using an unsupervised and data-driven approach. Originating from the concepts underpinning the Synaptic Kernel Adaptation Network (SKAN) \cite{Afshar2014c}\cite{afshar2015turn}, this method, called  the Feature Extraction with Adaptive Selection Thresholds (FEAST) algorithm makes use of neurons or features with individually adaptive selection thresholds that are iteratively updated using a competitive control strategy. These adaptive neurons act as feature extractors that learn data-specific features in an online, event-based, and unsupervised manner. Adaptive selection thresholds provide a simple way of maintaining homeostasis between the activation patterns of a large number of neurons without the need to store or share information about previous neuronal activity or the internal parameters of the individual neurons. This simplicity also enables efficient implementation of the algorithm in neuromorphic hardware. 

The unsupervised feature extraction algorithm is introduced and validated through classification tasks using both a controlled benchmarking dataset and a noisy real-world dataset. The choice of a classification task provides a quantitative and robust means of validating the information extracted by the feature detectors and the use of two different datasets demonstrates the applicability of the algorithm to a range of problem sets.

\subsection{Neuromorphic Vision Sensors}
\label{sec:intro-sensors}

Perhaps the greatest successes of neuromorphic engineering have occurred in the development of sensors that emulate biological perception of sensory transduction, with a very strong focus the development of novel and biologically-inspired vision sensors. By drawing inspiration from the functioning of biological retinas, neuromorphic vision sensors operate in a paradigm that is entirely different from conventional frame-based sensors  \cite{Roska2003}. Although the first electronic models of the retina were developed in the 1970's \cite{Fukushima1970}, it was the integrated silicon retina by Mahowald and Mead \cite{Mahowald1994} that represented the first viable neuromorphic imaging device. This device incorporated many of the characteristics found in the current generation of neuromorphic vision sensors, such as adaptive photoreceptors, spatial smoothing networks and an asynchronous communication paradigm.

These silicon retinas employ independent and asynchronous pixels that generate events only in response to changes in illumination at each pixel, producing data in an event-based manner and removing the need for frames and fixed exposure times. Since each pixel operates independently, the sensor as a whole can operate with a very high dynamic range and greatly reduces saturation and exposure effects. The data generated by event-based sensors has a very high temporal resolution allowing the sensor to capture high-speed activity in a sparse activity-driven manner.

This work makes use of the Asynchronous Time Imaging Sensor (ATIS) \cite{Posch2011}, which is a CMOS dynamic vision and image sensor with a resolution of $304 \times 240$ pixels, each containing a level-crossing detector (TD) circuit and an exposure measurement circuit (EM) for absolute illumination measurements.

As with many event-based neuromorphic vision sensors, the devices encode their output in the form of asynchronous spikes in the Address-Event Representation (AER) \cite{Boahen2000}. For the change detection circuit, this takes the form of events generated with a polarity to indicate an increase or decrease in the relative illumination, and for the EM circuitry, the absolute illumination level is encoded as the inter-spike interval between two separate events generated by the exposure measurement circuitry at each pixel.

\subsection{Feature Extraction in Neuromorphic Systems}
\label{sec:intro-feature-extraction}

Feature detection is commonly defined as the process of identifying and describing sections of an image representation for the purposes of identification, tracking, or classification. When dealing with conventional cameras, these image representations are often frames of illumination intensity, containing either monochrome or color information for each pixel. 

Feature detection in the context of event-based cameras operates on a fundamentally different representation of the visual scene in which the encoding of information includes a pixel-level temporal component not present in conventional frame-based data.

Event-based vision systems therefore need a class of feature detectors that exploit the event-based and activity-driven nature of neuromorphic vision systems. Tasks such as tracking and object recognition still require the identification and matching of local visual features, but these features must represent commonly observed spatiotemporal patterns of events instead of static images. As with conventional feature detection, the most desirable property of a feature is still its ability to be uniquely distinguished in feature space~\cite{Yang2011}.

The field of neuromorphic vision has seen a significant increase in interest over the past few years, resulting in a number of innovative approaches to the task of feature detection and extraction. Features based on corner detection, such as the Harris corner detector \cite{Harris1988} and cortex-like Gabor filters \cite{Kousa2007}, are examples of an explicit feature detection method commonly used in conventional computer vision. This approach has been adapted for event-based sensors, with notable examples including an event-based implementation of the Harris Corner detector~\cite{Vasco2016}, and a novel corner detection method based on finding the intersections of planes fitted to the event stream from the cameras \cite{Clady2015}. Related to corner detection is the process of edge detection, and this class of algorithms have also been implemented in an event-based manner. Examples include the Canny edge detector presented in \cite{Ieng2014} and the event-based line segment detector presented in \cite{Brandli2016}.

Whereas some feature detection methods have sought to make use of a combination of event-based and frame-based approaches~\cite{Tedaldi2016}, this work restricts itself to operating only on the output of the change detection circuity from the event-based camera. The mechanism for measuring absolute illumination in an event-based sensor varies from device to device, whereas the change detection produces compatible output across all current event-based vision sensors. By restricting the algorithms to only the change detection events, this maximizes the versatility and applicability of the algorithms by allowing them to be compatible with most existing event-based vision devices, such as the DAVIS event-based cameras~\cite{Brandli2014b}.

The HFIRST algorithm~\cite{Orchard2015} is an example of a multi-layer network in which appropriate features are learned directly from the event-based data. The algorithm is based on the HMAX algorithm~\cite{Serre2007} and implements an analogous first-to-spike operation in place of the maximum pooling operation from which the algorithm derives its name. Another example of an unsupervised learning method capable of learning spatiotemporal features makes use of recurrent reservoir networks and a winner-take-all approach~\cite{Lagorce2015} allowing the network to maintain and preserve the high temporal nature of the events throughout the feature detection system. The Event-Based GASSOM~\cite{Chandrapala2016} algorithm extends the ASSOM~\cite{Chandrapala2014} algorithm to the output of an event-based sensor and successfully demonstrated the ability to learn features invariant to fast changes in the input signal. Both methods were tested on event-based datasets similar to, or superseded by the datasets used to verify the feature extraction method presented in this work.

The Hierarchy of Event-based Time Surface (HOTS) algorithm~\cite{Lagorce2016} represents the closest work to the event-based feature detection method described in this work. The HOTS algorithm uses the neuron update learning rule introduced in ~\cite{ballard2012dynamic} where neurons are updated in proportion to the cosine distance of their weights to the input, and where the learning rate gradually decays as a function of time. The HOTS algorithm makes use of multiple layers of feature extractors based on unsupervised feature clustering with the output of each layer fanning out and feeding into deeper feature detectors with longer exponential time constants. The algorithm successfully demonstrates the ability to use these feature layers to perform accurate feature classification in an entirely event-based manner. 

This paper highlights the significance and benefits of using an adaptive thresholding approach to feature extraction as well as highlighting the novel use of readily accessible network signals for estimating weight convergence and predicting network classification performance. The adaptive thresholding technique presented allows a simple yet robust implementation of network homeostasis. Unlike in the learning method proposed in ~\cite{ballard2012dynamic} and used in HOTS, the neurons or features in this work do not need to continuously keep count of their updates. Instead the FEAST algorithm adapts its selection thresholds for incoming events such that the features are constantly being contracted by accepted events and expanded by rejected or missed events. This event-based competition means that the neuron thresholds do not decay exponentially as a function of time but only in response to missed input data which represents information not yet well incorporated into the network. 

Furthermore whereas the HOTS learning algorithm updates every neuron in proportion to the cosine distance of the input to the neuron, in FEAST only the winning neuron is updated. Designating a single neuron as the winner of an input event, and only updating this winner, not only performs a max pooling operation, but also significantly reduces computation and potential hardware costs of such an online learning system.

By using adaptive selection thresholds, the proposed network provides reliable signals for the detection of network convergence without requiring access to the network weights. The most direct indication of network convergence is a stable steady state in the values of the network weights. In general the change in the network weights does not reach zero but a steady state around a small value depending on the magnitude of learning rate. In this work three additional signals, the selection thresholds, the missed spike rate, and the variance in the output spike rates, are shown to provide alternative signals for convergence detection during online learning. This useful property is the direct consequence of the dynamics of the adaptive selection thresholds. Such proxy signals for weight convergence are unnecessary when a network is trained offline or if network weights are readily available for inspection and convergence analysis. However, in neuromorphic hardware applications, continuous access to network weights may be limited and costly, due to the large number of weights and limited number of output channels. The proxy signals examined in this work are more accessible and easier to calculate than the change in network weights, enabling more efficient implementations of neuromorphic on-chip learning hardware.

Thus the proposed algorithm trades a small amount of information loss for a simpler implementation of network homeostasis and robust measures of fitness to data and early proxies for classification accuracy which are shown to predict classification accuracy directly. These novel uses of intermediary signals are particularly important in the context of on-chip event-based neuromorphic systems in real-world online learning applications where hardware resources and opportunities for detailed network investigation are limited.

\section{Adaptive Threshold Clustering}
\label{sec:feast}

In this section the prerequisite event processing that precedes the feature extraction algorithm is described.

The output of an event-based camera can be viewed as a continuous stream of events $e$, each of which have the following form:

\begin{equation}
  \label{eq:event-definition}
  \mathbf{e_i} = [\mathbf{u_i}, t_i, p_i]^T  \quad i \in \mathbb{N}^+
\end{equation}

in which $ \mathbf{u_i} = [x_i, y_i]$ denotes the location of the pixel generating the event, $p \in [-1,+1]$ indicates the polarity of the change in illumination at the pixel causing the event, and $t$ represents the time at which the event occurred. In a hardware event-based system, the time-stamp would not need to be explicitly stored for each event, as the time would be implicit as the arrival time of the event during processing. Each event from the sensor has a time-stamp explicitly applied as it is retrieved from the camera. This time-stamp is stored with microsecond resolution.

Adopting a similar notation to that used in \cite{Clady2015}, and making use of surfaces inspired and based on the time surfaces presented in the HOTS algorithm \cite{Lagorce2016}, we can then define the function $\Sigma_{e}$ to map a time $t$ to each 2D spatial coordinate $\mathbf{u}$:

\begin{eqnarray}\label{eq:sigmasurface}
  \Sigma_{e}:  &\mathbb{R}^{2} \to \mathbb{R} \nonumber \\
  \mathbf{u}: &{t} = {\Sigma_{e}(\mathbf{u})}
\end{eqnarray}

and a similar function $P_{e}$ to map the polarity to each spatial coordinate:
\begin{eqnarray}\label{eq:polaritysurface}
  P_{e}:  &\mathbb{R} \to \left\{ {-1,1} \right\} \nonumber \\
  \mathbf{u}: &{p} = {P_{e}(\mathbf{u})}
\end{eqnarray}

As time is inherently a monotonically increasing function, the function $\Sigma_{e}$ defined in \eqref{eq:sigmasurface} describes a monotonically increasing surface. Applying a function to this surface generates other surfaces from which more descriptive features that better represent the underlying visual scene can be extracted. Two potential surfaces $\Lambda_{e}(\mathbf{u},t)$ and $\Gamma_{e}(\mathbf{u},t)$ which are derived from $\Sigma_{e}$ are described below.

The simplest candidate surface, is the fixed time window time surface $\Lambda_{e}(\mathbf{u},t)$, which is defined as:

\begin{equation}
  \label{eq:expdecayingtimesurface}
  \Lambda_{e}(\mathbf{u},t) = \begin{cases}
    P_{e}(\mathbf{u}),
    & 0 \leq t - \Sigma_{e}(\mathbf{u}) \leq \tau \\
    0,
    &  \tau < t - \Sigma_{e}(\mathbf{u}) < 0\\
  \end{cases}
\end{equation}

The above equation utilizes the two functions defined in equations \eqref{eq:sigmasurface} and \eqref{eq:polaritysurface} to create a trinary valued surface where the $\tau$ parameter defines the duration over which an event will have a non-zero value on the surface and therefore is responsible for implementing the required memory aspect needed for spatiotemporal pattern identification. The actual value for $\tau$ is dependent on the application and nature of the visual field and needs to be specified whenever such a surface is used. 

A second more complex candidate surface extends this concept of the time surface by implementing an exponential decay rather than a fixed time window. In this way the weight of the information carried by each event is decayed smoothly toward zero over time. The exponentially decaying time surface $\Gamma_{e}(\mathbf{u},t)$ is defined as follows:

\begin{equation}
  \label{eq:expdecayingtimesurface}
  \Gamma_{e}(\mathbf{u},t) = \begin{cases}
    P_{e}(\mathbf{u}).\mathrm{e}^{\left(\frac{\Sigma_{e}(\mathbf{u}) - t}{\tau}  \right)},
    & \Sigma_{e}(\mathbf{u}) \leq t \\
    0,
    & \Sigma_{e}(\mathbf{u}) > t
  \end{cases}
\end{equation}

As with the previous surface, the $\tau$ constant in the equation determines the duration over which events have impact on the scene. However here $\tau$ is defined as the decay constant of the exponentially decaying surface activation due to incoming events. A $\tau of 10 ms$ and $\tau of 10 ms$ were used for the N-MNIST dataset and Plane dropping dataset respectively.

\subsection{Feature Detection from Time Surfaces}
\label{sec:feature-detection}
\begin{figure*}
  \centering
  \includegraphics[width=\textwidth]{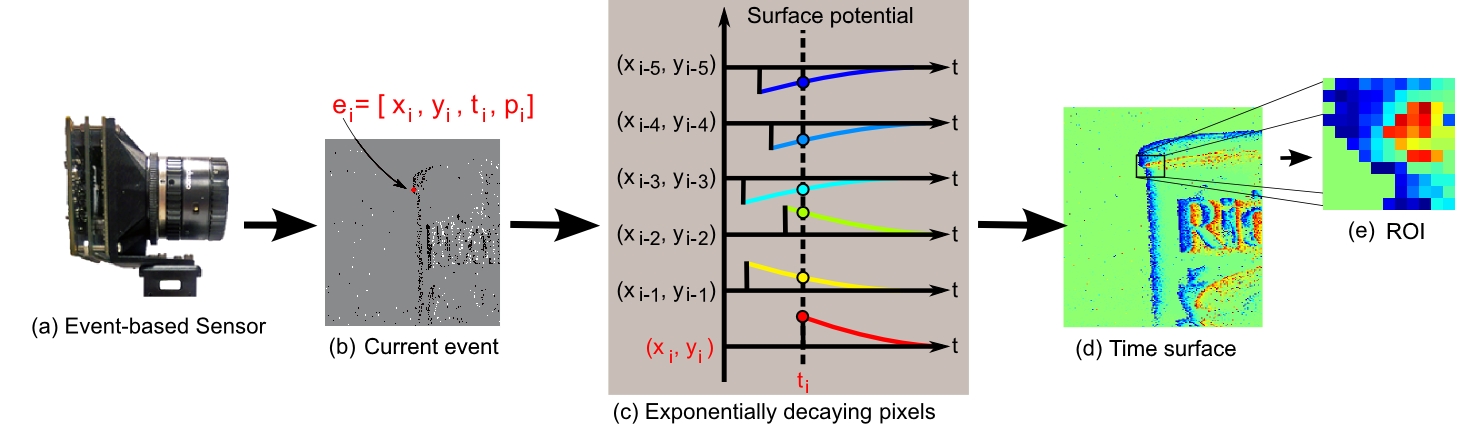}
  \caption{\textbf{Construction of a spatiotemporal time surface and features from the event-based output of an ATIS camera.} The time surface encodes the temporal information from the sensor into a continuous value on the surface in response to the arrival of each event. The ATIS sensor shown in (a) generates a stream of events, with an example of a single event shown in (b). When an event occurs, such as event $e$ from the pixel $(x_{i}, y_{i})$ at time $t_{i}$, the value of the surface at that point is changed to reflect the polarity $p \to \left\{ {-1,1} \right\}$. An example of the decaying exponentials that generate the time surface at $ t = t_{i}$ is shown in (c). The exponentially decaying time surface is shown in (d). The intensity of the colour on the surface encodes the time elapsed since the arrival of the last event from that pixel. The input to the feature extractor in the Region of Interest (ROI) which consists of a $w \times w$ window centered on the event $e$, as shown in (e).}
  \label{fig:time-surface-diagram}
\end{figure*}

An event-based approach introduces the challenge of detecting desirable features in a spatiotemporal data stream, and the challenge of not only detecting where points of interest occur, but also when they occur. Figure~\ref{fig:time-surface-diagram} shows an illustration of the event context extraction from time surfaces in response to an incoming event $e$. The use of a time surface reduces the data stream into a two dimensional representation making it possible to generate frames similar to those shown in Figure~\ref{fig:time-surface-diagram}b. Such frames can be generated at regular intervals from these surfaces allowing the application of conventional feature detection techniques. However, such a frame-based approach would discard the high temporal resolution offered by the sensor and potentially result in non-optimal feature-sets.

Instead, the derived event time surfaces shown in \ref{fig:time-surface-diagram} can be used to extract a Region of Interest (ROI) patch around every single incoming event. Since it is not feasible to process the entire image resolution for each incoming event, limiting the feature information to the local spatial ROI around the event allows a computationally efficient event-based input to the feature extraction algorithm.

Extracting an event ROI from the time surface for an incoming event produces an $w\times w$ ROI $I_{e}$ containing spatiotemporal information from the  neighborhood surrounding the pixel generating the event. In this work the ROI patch size was selected as the neighboring $11\times11$ pixels. In order to perform further processing on this event, the ROI region is converted into a descriptor in the form of a one dimensional, $1\times w^2$ vector as follows:
\begin{equation}
  \label{eq:descriptor-vectorization}
  d = vec(I) = [I_{1,1}...I_{w,1}, I_{1,2}...I_{w,2}, I_{1,w}...I_{w,w}]^T
\end{equation}

In the next step this descriptor is normalized through a division by its norm to achieve invariance to temporal scaling.
\begin{equation}
  \label{eq:descriptor-normalised}
  d = \frac{vec(I)}{||vec(I)||}
\end{equation}

The time scale invariance resulting from normalizing the descriptor is an important operation in the weight update step of the FEAST algorithm. Since the magnitude of the values contained in $I$ encode the last spike times for each pixel in the ROI, the relative difference between pixels in $I$ is effectively a measure of scene velocity (generated either by moving objects within the scene, or of the camera itself). This measure of local velocity allows the calculation of optical flow about an event\cite{Benosman2014}. By effectively normalizing the descriptor $I$ with respect to time, the velocity information is discarded in favor of feature robustness.

\subsection{Adaptive Threshold Clustering}
\label{sec:adaptive-threshold-clustering}

An event-based system requires an online method of determining the optimal prototypical features for the dataset, and the k-means clustering algorithm \cite{MacQueen1967} is perhaps a logical candidate for such an operation. Indeed, k-means clustering allows for a preselected number of features to be iteratively computed on the arrival of new data. 

Unfortunately, the k-means clustering algorithm is not a perfect candidate for clustering on event-based data. It is sensitive to the initial estimates for the features \cite{Likas2003}, and the choice of initial estimates in a high dimensional space is not a trivial problem.

The choice of update rule for k-means clustering is also a concern, as there are two main varieties tailored to handling different situations. The first update rule weights each incoming sample by the total number of samples received. This version allows for features that encompass a large dataset to be iteratively determined without the loss of any information. Unfortunately, such an approach places emphasis on the initial training sequences, and is poorly suited to a continuous operation (as required by an event-based system) as each subsequent event exerts less influence on the feature weights.

The second update rule uses a fixed threshold by which to update the features. This implements a system similar to a low-pass filter, and allows the features to continuously evolve to match the incoming data, making it a better match for an event-based system. 

Here we introduce a simple means of performing online unsupervised feature extraction using a method of clustering with an adaptive selection threshold for each feature. Under this setup, each feature possesses a unique adaptive selection threshold which encodes the minimum permissible acceptance distance between the feature and a new incoming descriptor. This threshold is dynamic, and given $i$ features, the thresholds change based on two rules:

\begin{enumerate}
 \item If the input ROI matches any feature (the cosine distance between its weights and the input ROI is within the feature's threshold), then the threshold $t_{i}$ is decreased for feature $i$ by a fixed amount $\Delta I$. (If multiple features match the input, the best matching feature is selected).
 \item If an incoming ROI does not match any of the features, then all thresholds are increased by a fixed amount $\Delta E$.
\end{enumerate}

\begin{figure}
  \centering
  \includegraphics[width=0.45\textwidth]{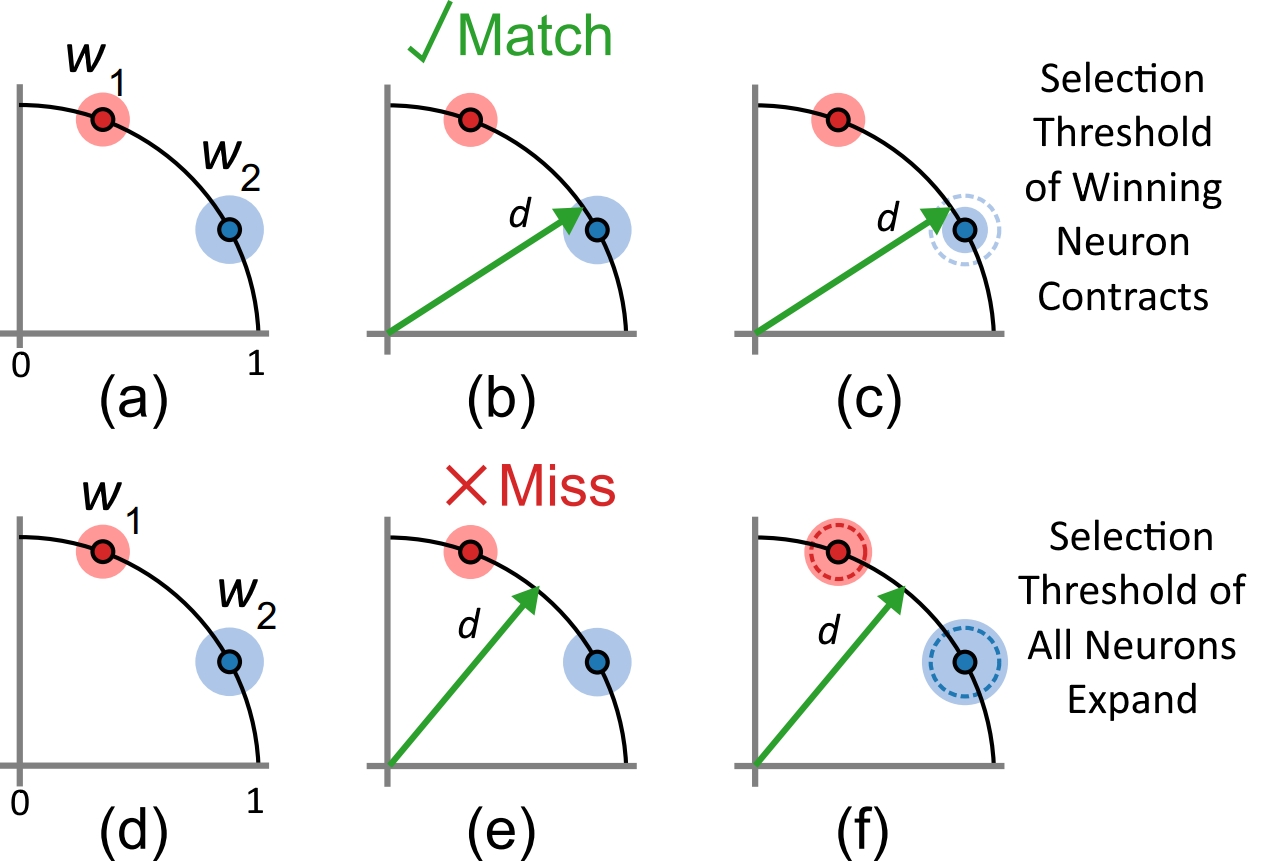}
  \caption{\textbf{Diagram showing the adaptive threshold clustering in a 2D space.} The diagram above shows a simplified case of the adaptive threshold clustering applied to features consisting of two dimensions. An existing configuration with two normalized features is shown in (a) on the unit circle, and with independently configured thresholds. If a following input ROI falls within the selective threshold distance of an existing feature, as shown in (b), then it is said to have matched and is assigned to the that feature. The threshold for that feature is then reduced as shown in (c). If the event does not match any existing features, as in (e), the input is discarded (missed event) and the thresholds for all features are increased as in (f).}
  \label{fig:fspace-surface-thresholdDiagram}
\end{figure}

As illustrated in Figure~\ref{fig:fspace-surface-thresholdDiagram} the normalization of the incoming ROI and the initialization of the feature weight vectors together with the unit gain of Equation~\ref{eq:descriptor-normalised} ensures that all are represented as points on the unit hypersphere, whose dimensionality is equal to the number of pixel in the ROI. 

Figure~\ref{fig:fspace-surface-thresholdDiagram} shows an illustration of the adaptive threshold operation as applied to features containing only two dimensions. In practise, the number of dimensions is far larger. The figure depicts two existing features $i_1$ and $i_2$, both of which have independent thresholds indicated by the spheres surrounding them\footnote{In reality, the normalization ensures that input falls on the unit circle in 2D, and therefore the matching region is actually an arc on the unit circle rather than a sphere, but is drawn in the figure as such for illustrative purposes.}. 

The distance between the input ROI and the current state of each feature is calculated using the cosine distance measure. Of all the features for which the ROI is within threshold, the feature with the smallest distance to the ROI is chosen as the matching feature. If the ROI does not land within the selection threshold of any of the features, no feature is chosen and no feature event is generated. If successfully matched to a feature, the input is then assigned to it, and the feature generates a feature event which can be viewed as an output spike from that feature neuron. 

When a feature is matched, the input ROI is incorporated into that feature by updating the feature weights with a fixed mixing rate $\eta$ as follows:
\begin{equation}
  \label{eq:eta-update}
 w_{n} = (1 - \eta) w_{n} + \eta d
\end{equation}
in which $w_{n}$ denotes feature $n$ to which the input ROI descriptor $d$ is successfully matched. As a point of reference, the mixing rate used to update the features in this work was set as $\eta = 0.001$.

The normalization of the descriptors described in Equation~\ref{eq:descriptor-normalised} means the input descriptor $d$ generated from each event injects an equal amount of information into the network by ~\ref{eq:eta-update}. Without the normalization step, faster moving segments of the scene, which have higher magnitude, would have a larger effect on the learned feature weights, in comparison to slow moving features. This in turn would result in feature sets that were biased toward faster moving objects.

In addition to updating the feature weights, the threshold for the winning feature is also decreased by $\Delta I$ as shown in Figure~\ref{fig:fspace-surface-thresholdDiagram}c. This contraction of the selective threshold of the winning feature, slightly reduces the receptivity of the feature to new inputs forcing it to become more selective with each win.

If the input does not match any existing feature, as in Figure~\ref{fig:fspace-surface-thresholdDiagram}e, the input is discarded and the thresholds for all features is increased by $\Delta E$. This has the effect of increasing the receptivity of all features making them more willing to accept input ROIs with greater distance to their current location in the feature space. Thus, with each 'missed' input event, the network as a whole becomes less selective and more receptive to change.

If the features are coding poorly for the incoming data (if, for instance, some subset of the features have become too specialized to an uncommonly observed set of points in the input space), then the global threshold increase serves to expand the range of input ROIs to which all features will respond. This increase in receptivity of the network will continue until the input data falls within the threshold of some of the neurons.

The effect of this dynamic thresholding serves to ensure that after convergence the rate of firing of for all features is approximately equal, as decreasing the threshold on matching serves to reduce the receptiveness of each feature to new data. This balance between expansion and contraction of the thresholds results in features with weights placed at the center of the mostly commonly observed regions of the input space, with selective thresholds that match the dispersion of observed inputs around those points. This adaptive method takes advantage of the abundance and informational redundancy in event-based data. By trading off a small fraction of events which are missed by all neurons, the algorithm finds the appropriate set of selective thresholds which balances the feature activation over the dataset.

\begin{figure}
  \centering
  \includegraphics[width=0.5\textwidth]{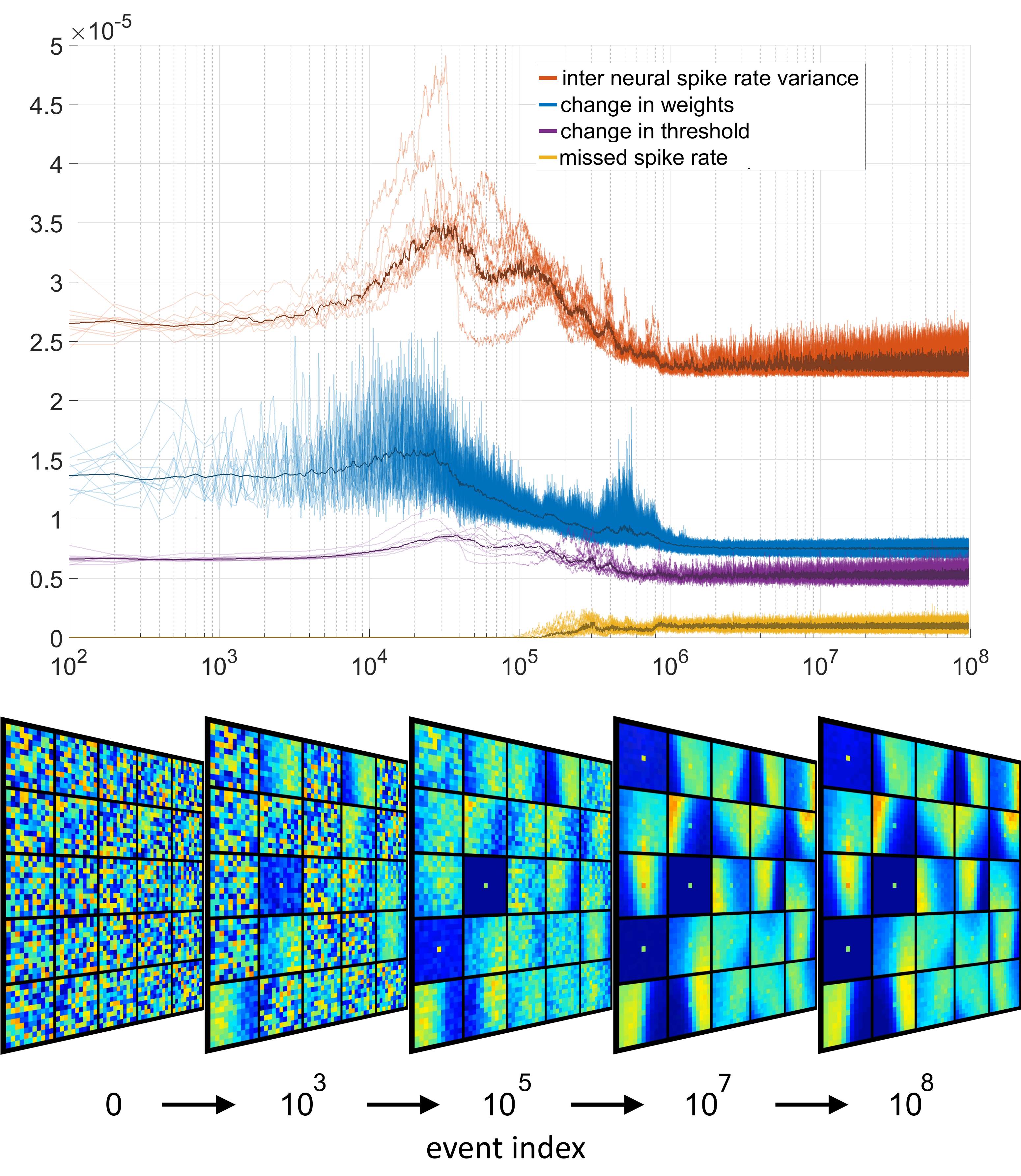}
  \caption{\textbf{Evolution of the network variables.} Top panel shows the adaptation of various neural signals in the network over 10 independent trials as well as the mean over the trials. Every 100 events four signals were sampled and and the magnitude of their inter-sample change is plotted over input event index. The signals plotted include the variance (across neurons) of the output spike rate (divided by ten thousand), the magnitude of the change in the synaptic weights of all neurons, the magnitude of the change of the selection thresholds and the missed spike rate (divided by ten thousand). The bottom panel shows the evolution of the feature weights for one of the trials. Note the presence of three variants of the "noise feature" in the feature sets. These empty features with a single high value at the central triggering pixel are learnt from events which are not correlated with any recent adjacent events. While the features appear empty and flat except for the central pixel, all but one feature per network typically exhibit very weak structure in their "empty" regions distinguishing them from each other and the dominant truly flat noise event}
  \label{fig:fspace-feature-generation}
\end{figure}

As shown in Figure~\ref{fig:fspace-feature-generation}, at the beginning of training the features are initialized to random points on the unit hyper sphere. Since the selection thresholds are initialized at random, a minority of the neurons will have thresholds so wide that every input event causes a neuron to fire, preventing other neurons from spiking and thus learning. Because of their greater receptivity, these neurons capture all input events such that there are no missed spikes and no change in the selection thresholds of the more selectively initialized neurons. This is evidenced by the top panel of Figure~\ref{fig:fspace-feature-generation} where during the initial stage the relative magnitude of change in the thresholds is low since only the thresholds of a few neurons are adapting (becoming ever more selective).

The magnitude of the change in the feature weights is similarly low due to the early unbalanced activity of the network, allowing only a few neurons to learn. An additional signal shown in the panel is the standard deviation of spike rates across neurons. At the early stage of learning, with only a few highly receptive neurons firing, the variance in the firing rates across neurons the neurons is low.

As the early highly receptive neurons contract their thresholds and become more selective, more neurons with more selectively initialized thresholds become activated and begin learning. This learning results in an increasing rate of change in the feature weights and the selection thresholds. Even greater change is observed in the variance of the output spike rate across neurons, as more and more neurons become activated while the most selective neurons have still not fired a single output. Eventually, as the number of activated neurons with decreased thresholds increases to a tipping point, the magnitude of the change in thresholds begins to decline as fewer and fewer highly receptive neurons are left for adaptation. Simultaneously, the change in weights and the variance in the spike rate of the neurons also falls, as the neuron weights and thresholds begin to take on the statistics of the input dataset such that the neurons orient toward the centroids of the most common spatiotemporal pattern clusters while the thresholds take on values in proportion to the spread of the patterns around these centroids. Eventually all neurons become so selective that some input events start to fall outside the selection threshold, causing the first missed spikes. With these missed spikes the thresholds of all neurons increases causing the final most selectively initialized neurons to respond to input and begin adapting their weights. After this point all signals move toward their final steady state values demonstrating the convergence of the network. In this state the change in weights and thresholds and inter neuron spike variance reach their lowest value, while the missed spike rate reaches a steady state of approximately 2 percent.

Once training is completed the selection threshold can be discarded such that during inference, the feature with the smallest cosine distance to the input is assigned to the incoming event, regardless of the absolute value of the adapted selection threshold. 

This adaptive selection threshold clustering maps the spatially encoded events of the sensor to the feature space of the network. The output of the network is in the form of $e_{j} = [F, t]^{T}$, where $F \in [F_{0}...F_{N}, N < b]$ corresponds to the feature (out of $b$ features) matched to the current event, and $t$ represents the time at which the original event occurred, i.e., it remains unchanged from the input. Note that in the FEAST method unlike that in HOTS the index $j is not equal to $i and is smaller by a small fraction equal to the number of missed spikes.

Examining the nature of the output, which now only contains a feature number $F$ and a timestamp $t$, the benefits of operating in feature space become apparent as the size of the output space (the number of features) is significantly smaller than the input space (number of pixels in the sensor).

\subsection{Noise features and network size selection}
A heuristic developed during the testing of the FEAST algorithm was to use the number of noise features to select the appropriate network size. Optimal feature receptive field size and the corresponding layer size are among the most difficult event-based network meta parameters to optimize. This is due to the large potential parameter search space, the strong interdependence of the parameters, and the long feedback loop guiding the parameter selection. For multi-layer networks the search space increases in a combinatorial manner. Furthermore, each data point in the network structure search space requires development and convergence of multiple independent feature extractor networks and subsequent multiple classification operations. 

To bypass this search we use a heuristic method of observing the noise features shown in Figure~\ref{fig:fspace-feature-generation} for selecting network size. These noise features are often one of the dominant features in networks trained on noisy real world event-based dataset. By representing noise events that are triggered by the noise in the sensor and not by changes in illumination, such noise features effectively perform unsupervised noise filtering. In addition, in this work these noise features are used for the novel purpose of selecting the number of neurons in each layer for a given dataset. This method is based on the observation that irrespective of the complexity and structure of the dataset the input descriptors $d$ generated by noise events are highly correlated with each other and contain mostly redundant information. This would ideally be described by a single noise feature neuron with all other neurons coding for the complex structured features present in the dataset. Thus in an extreme case, a large network whose trained features are all variants of the noise feature is evidence of a training dataset that contains no structured information beyond noise events. At the other extreme, again assuming a non-ideal sensor which generates some noise, a network containing only complex features with no noise features is evidence of a network containing too few neurons since it has yet to incorporate the information from the noise feature (and presumably, other more significant non-noise features also). In this heuristic approach, if an event-based sensor is assumed to generate noise events, the target number of noise features learnt from any dataset should be at least one and possibly slightly more (to ensure that non-noise features less common than the noise feature are also incorporated). 

In practice, as the number of neurons in the feature set is increased and representation of unique complex structured features in a dataset is exhausted, the number of additional noise-like features tends to increase. After this point, given that noise features by definition do not correlate to any target class, further increase in neuron numbers is likely to produce diminishing returns. In this work a target of 2-4 noise features was selected for both the N-MNIST and Plane Dropping dataset. This target resulted in respective layer sizes of 100 and 25 neurons per polarity for the he N-MNIST and Plane Dropping datasets respectively. This heuristic method of selecting layer sizes by observing the number of noise features becomes particularly important for feature extraction networks developed for noisy real-world event-based data and in applications where exhaustive search of network structures is unavailable.

\section{Methodology}
\label{sec:methodology}
This section presents a brief introduction to the datasets and methodology used to perform the event-based adaptive threshold clustering required for the FEAST algorithm. The two datasets used to test and verify the work are introduced and a brief description of the means by which time surfaces are used to generate spatiotemporal event contexts are presented. This is followed by a description of the event-based feature extraction method and discusses the ability for these methods to act as a noise filter. Finally, the classifiers used to perform the feature detection are also introduced and in this section.

\subsection{Datasets}
\label{sec:datasets}
Two different datasets are used to explore and a validate the FEAST algorithm. There are a growing number of standardized event-based datasets captured with event-based cameras. These include datasets for a wide range of vision-related tasks, such as action recognition~\cite{Hu2016}, optical flow~\cite{Rueckauer2016}, face recognition~\cite{Lagorce2016}, and visual navigation~\cite{Barranco2016}. 

\begin{figure}
  \centering
  \includegraphics[width=0.3\textwidth]{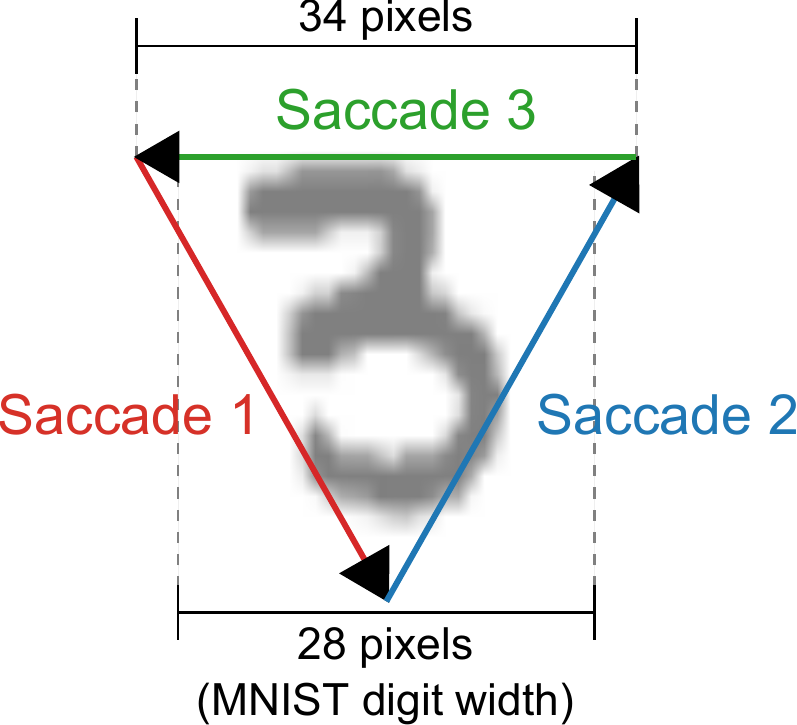}
  \caption{\textbf{Diagram showing the three saccade-inspired motions across each digit used in creating the spiking neuromorphic datasets.} Each MNIST digit is sized so that it occupies a $28 \times 28$ pixel region on the ATIS sensor, and three saccade motions starting in the top left and moving in triangular motion across each digit. Due to the motion, the resulting digits sequences span a $34 \times 34$ pixel region to ensure that the full digit remains in view during the entire saccade motion.}
  \label{fig:recog-detect-saccadediagram}
\end{figure}

The adaptive threshold clustering algorithm was tested on the N-MNIST event-based dataset~\cite{Orchard2015b} to present and characterize the feature extraction process. The N-MNIST dataset is a conversion of the original and widely disseminated MNIST dataset~\cite{LeCun1998} to an event-based format. Whereas the letter and digit dataset presented alongside the HFIRST algorithm contained only a small subset of digits, the N-MNIST dataset contains the full 70,000 training and testing samples. The dataset is converted to an event-based representation by projecting the digits onto an LCD screen and then recording them with an ATIS camera as it proceeded through three defined saccade-like movements forming the triangle shown in Figure~\ref{fig:recog-detect-saccadediagram}.

While the N-MNIST dataset represents a good benchmarking task for event-based classification systems, with results reported in multiple papers \cite{Cohen2016,Chandrapala2016,JLee2016,cohen2018spatial}, it represents a heavily controlled classification task. The use of the fixed and predictable saccade motion creates an unrealistic assumption on which to test feature detection algorithms destined for less controlled real-world tasks. In order to better evaluate the performance of the feature detectors, a less controlled classification task was added which introduces more noise and greater natural variation into the task whilst still providing well-labeled ground-truth data.

\begin{figure}
  \centering
  \includegraphics[width=0.45\textwidth]{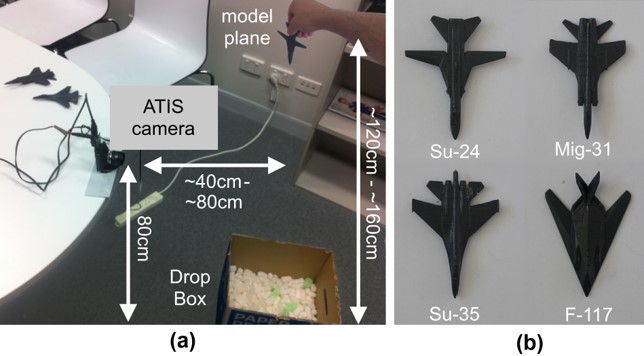}
  \caption{\textbf{The setup for the collection and creation of the plane dropping event-based dataset.} The figure shows a photograph illustrating the means by which the Plane Dropping dataset was collected. The annotated photograph shown in (a) shows the ATIS camera attached to a table at a fixed height of 80~cm from the ground whilst small model planes are dropped freehand from a height of between 120~cm and 160~cm in front of the camera. The four model planes, shown in (b), are approximately 10~cm in length and were all painted uniform gray to remove any textures or markings on their surface. It should be noted that no effort was made to control the lighting in the room, nor the exact height, position, or angle at which the planes were dropped. }
  \label{fig:plane-drop-dataset}
\end{figure}

The task in this dataset is the identification of small model planes dropped from standing height and captured as they rapidly passed the field of view of the event-based camera in an average of 240~ms. Figure~\ref{fig:plane-drop-dataset} provides an annotated photograph of the experimental configuration and shows the four model planes used to create the dataset. Each plane was dropped 100 times freehand without any means to regulate the height or angle at which the planes were dropped.\cite{afshar2018investigation}

The Plane Dropping dataset provides a more realistic dataset for event-based classification, as it represents a tasks that highlights the benefits of the event-based cameras and is difficult to replicate using conventional sensors. The high-speed nature of the task would require the use of high-speed frame-based cameras that produce data at a significantly higher data rate than the event-based sensors, highlighting the benefits of the sparse change-based output of the neuromorphic sensing approach. 

The dataset consists of 100 drops of each airplane type and contains significant variability in terms of speed, orientation, and position relative to the camera. Additionally, there are variable delays before and after each drop, resulting in recordings of varying lengths. This dataset was additionally augmented with left-right flipped versions of the recordings resulting in 200 drops for each airplane type. The generated dataset is constrained in the sense of having a single high speed object in the field of view against a background that generates few events. This restriction allows an efficient, focused investigation of the sources of variance in the airplane dataset such as noise, target orientation, and velocity. In this way the source of variance in the two tested datasets can be viewed as complementary. 

\subsection{Classifiers}
\label{sec:classifiers}

Two different classifiers were used to perform the learning and classification tasks on the feature events generated from the FEAST algorithm. The first classifier is an iterative implementation of the Extreme Learning Machine (ELM)~\cite{Huang2006}. ELM networks consist of a standard three layer configuration and use random weights to project from the input layer to a hidden layer. This hidden layer input is passed through a nonlinear activation function, typically a sigmoid function. A set of linear output weights are then learned to map the hidden layer output to the output classes, thus performing classification. 

The classifier uses the Online Pseudo-inverse Update Method (OPIUM)~\cite{VanSchaik2015b} to iteratively update the linear output weights which project from the hidden layer neurons to the output neurons. The use of an iterative method of solving the pseudo-inverse for the ELM allows the classification network to be updated in response to each individual event, however, the scale of the number of input channels and the size of the event-based dataset make direct application of the ELM network to the event-based data prohibitively difficult, motivating the dimensionality reduction provided by the feature extractor network.

The second classifier used in this work is the Synaptic Kernel Inverse Method (SKIM)~\cite{Tapson2013a}, which is a neural synthesis technique designed to operate directly on spike-based inputs and is therefore directly compatible with the event-based output of these event-based sensors. The SKIM method is inspired by the process of dendritic computation and has a similar three-layer structure to the ELM network with fixed and random connections from the input layer to a much larger hidden layer. A set of learned linear weights connect the hidden and the output layer neurons. The hidden layer neurons in SKIM represent dendritic synapses and implement a nonlinear activation function, whereas the random input weights model the axonal connections to other neurons. The linear output weights represent the dendritic connections to each output neuron.

The SKIM network also makes use of OPIUM to learn these weights, although any gradient descent method would be suitable. The original SKIM network as proposed by Tapson et al. provided a means of synthesizing networks capable of producing specific spatiotemporal patterns in response to specific input patterns. The implementation of the network requires a number of modifications in order to utilize the algorithm for a classification task. A full discussion of these alterations is provided in \cite{Cohen2016}.

The two major alterations involve the nature of the training signal and the means by which the classification output is determined in a multi-class classification task. The original SKIM network made use of a single spike as the training target, which performed well on simple spike pattern recognition tasks but did not extend well to larger and more complex datasets, suffering from a high degree of sensitivity to noise.

These single output spikes were replaced with a multi-step training signal which imparts more energy into the system and greatly increases the robustness of the learning mechanism. A range of possible transfer functions can be used for spreading the energy of the events to later training time steps. These include linearly decaying, exponentially decaying or Gaussian transfer functions. For this purpose the SKIM classifiers presented in this use Gaussian transfer functions were used as training signal which was amended to the end of each recording. A second adaptation required to handle multi-class classification tasks involves the means by which the winning class is determined. Whereas in ELM networks, the class with the maximum activation at each time step is the winning output class, this assumption does not readily translate to the event-based paradigm used in the SKIM network. In this work, additional time-steps are appended to the end of the training and testing spike trains. During this augmented section at the end of the spatiotemporal pattern, the supervisory signal indicating the winning class is activated. For the purposes of this work, the winning class is determined by the output neuron with the highest cumulative activation during the augmented period. This is referred to as the Area determination method, as described in \cite{Cohen2016}.

\section{Results}
\label{sec:results}

\subsection{N-MNIST Digit Classification Results}
\label{sec:n-mnist-digit}

For the purposes of the classification experiments, the FEAST method was used to generate features on the N-MNIST dataset. Only the training samples were used to generate the features, and made use of the feature extraction parameters configured as $\Delta I = 0.001$ and a $\Delta E = 0.003$ with 200 features (100 for ON events, 100 for OFF events) of size $11 \times 11$ pixels.

Through the heuristic examination described in section~\ref{sec:Noise features and network size selection} N-MNIST dataset, 100 features were selected as the feature layer size for each polarity, resulting in a network containing a small number of noise features.

\begin{figure}
  \centering
  \begin{tabular}{c|c}
    \subfigure[]{\includegraphics[width=0.45\columnwidth]{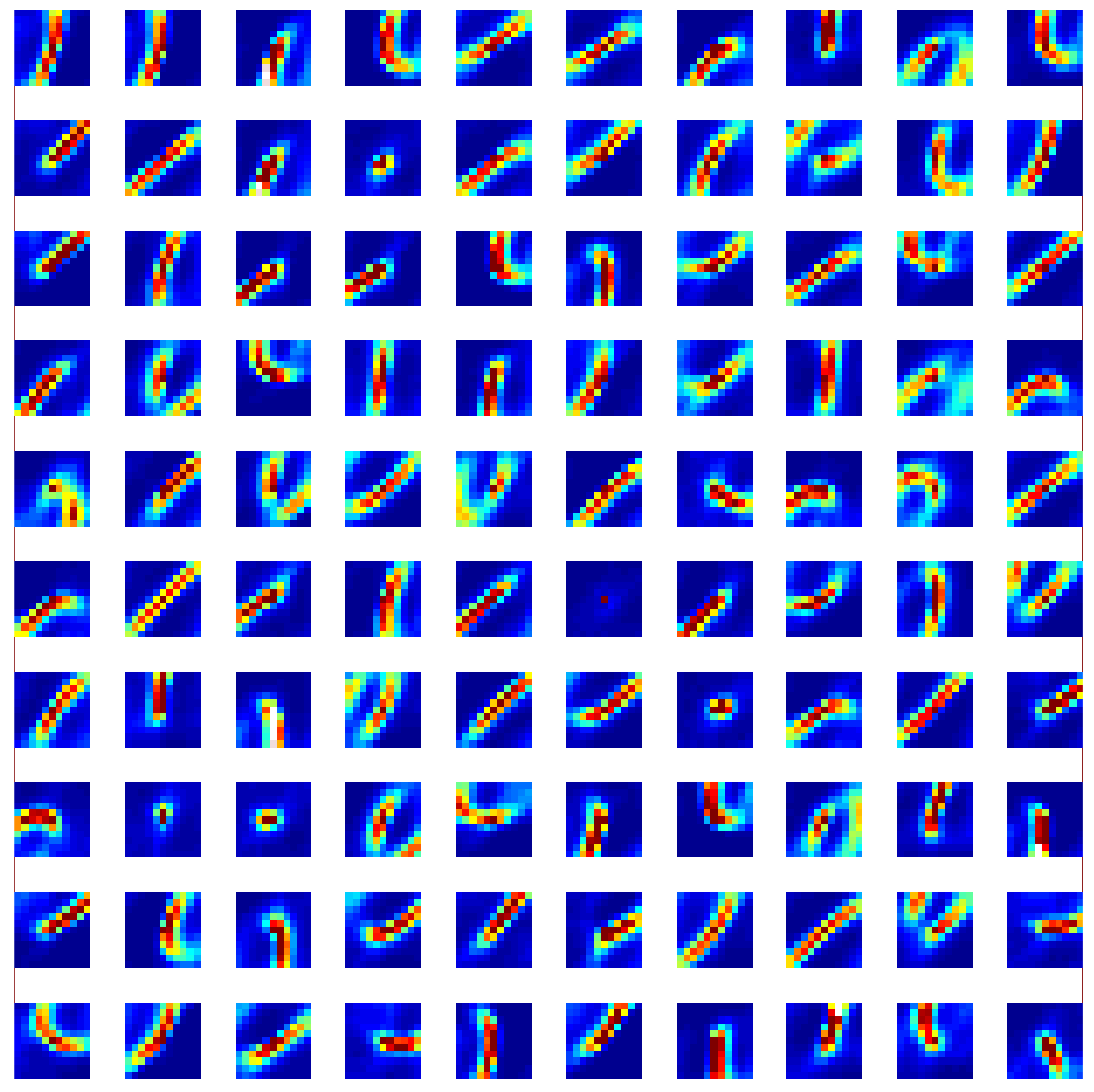}}
    &
    \subfigure[]{\includegraphics[width=0.45\columnwidth]{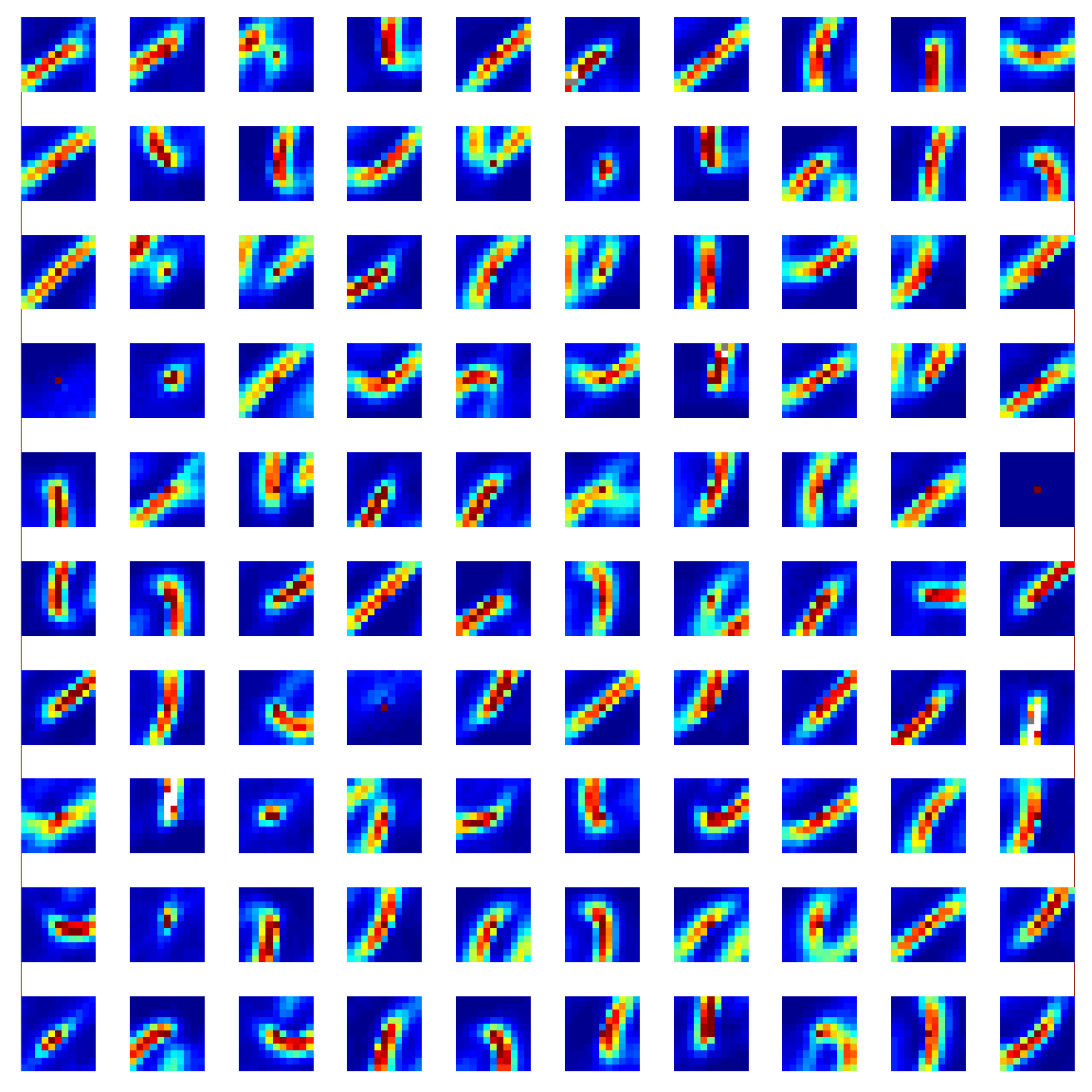}}
  \end{tabular}
  \caption{\textbf{$11 \times 11$ features learned from the ON events (a) and off events (b) of the N-MNIST dataset.} Each feature represents a normalized vector reshaped to match the size of the incoming feature patches}
  \label{fig:fspace-surface-100features}
\end{figure}

Figure~\ref{fig:fspace-surface-100features} shows the resulting 200 features of $11 \times 11$ image patches from the N-MNIST dataset.

An ELM network was used to verify that the extracted features provide a suitable basis with which to perform classification. ELM networks do not intrinsically operate on event-based data, and therefore the input sequences from the N-MNIST dataset cannot serve directly as input for an ELM network. The most direct approach requires the use of a separate input channel for each pattern at each time-step, and when dealing with the original N-MNIST sequences, the image size of $34 \times 34$ pixels and the 316 time-steps (the maximum number of millisecond time-steps in the N-MNIST dataset) results in a required input size of 365,296 per digit which is prohibitively large. However, by mapping the data into the feature domain, the size of the input layer is reduced to a single feature per time-step. For a network containing 100 features, this results in an input pattern size of 31,600, reducing the input layer by more than an order of magnitude. 

Testing with the ELM classifier involved the two different sets of features for each polarity, with the same event to feature mapping method used for each polarity. 

\begin{table*}[htbp]
 \caption{\textbf{Classification accuracy for the ELM  network on raw events and on FEAST events and the SKIM network using random and FEAST features.} The classification accuracy is reported in terms of percentage of digits correctly identified using 200  $11 \times 11$ pixel features. Ten trials of each configuration were performed. }

 \begin{center}
  \begin{tabular}{|l|c|c|c|c|c|c|c|c|}
  \hline
  {} & \multicolumn{2}{|c|}{ELM + Random Features} &   
  \multicolumn{2}{|c|}{ELM + FEAST Features }  & 
  \multicolumn{2}{|c|}{SKIM }  &
  \multicolumn{2}{|c|}{SKIM + FEAST } \\
  \cline{2-3}
  \cline{4-5}  
    \cline{6-7}
  \cline{8-9}  
  {Hidden Layer} & Mean & $\sigma$ & Mean & $\sigma$ & Mean & $\sigma$ & Mean & $\sigma$ \\
  \hline
1000 Neurons  & 75.53  \%   &  0.28 \%   & 84.93 \%  & 0.31 \%     &81.75\% &0.38\%   &85.97\%&0.69\% \\
  2000 Neurons & 83.03 \%   &  0.25 \%  & 89.56 \%  & 0.28 \%   &86.80\%& 0.29\%      &90.47\%&0.45\%  \\
  3000 Neurons   & 86.37 \%   &  0.28 \%  &  91.45 \%  & 0.22 \%   &88.79\% &0.26\%      &91.88\%&0.34\%    \\
  4000 Neurons & 87.92 \%  & 0.24 \%   & 92.39 \%  & 0.18 \%     &89.97\% &0.19\%    & 92.75\%&0.21\% \\
  5000 Neurons & 89.01 \%  & 0.17 \%   & 92.87 \%  & 0.16 \%      &90.69\% &0.25\%   & 93.43\%&0.25\% \\
  6000 Neurons & 89.74  \%  & 0.23 \%   &  93.30 \%  & 0.20 \%     &91.25\% &0.21\%    &93.73\%&0.08\%  \\
  7000 Neurons & 90.27 \%  & 0.27 \%  & 94.06 \%  & 0.13 \%      &91.54\% &0.12\%   & 94.11\%&0.24\%     \\
  8000 Neurons & 90.50 \%  & 0.11 \%  &  94.00 \%  & 0.12 \%    &91.85\% &0.16\%     &94.25\%&0.15\%   \\

  \hline
  \end{tabular}
 \label{tab:surface-features-elm-results-summary}
 \end{center}
\end{table*}

Table~\ref{tab:surface-features-elm-results-summary} presents the results of the ELM classifier and SKIM networks. Eight different hidden layer sizes were tested, and ten trials of each experiment performed. The results show that the learnt FEAST features outperform the random features at every tested hidden layer size, and by a significant percentage, achieving an overall accuracy of 94\% with 8000 hidden layer neurons. As shown in Table~\ref{tab:surface-features-elm-results-summary}, the standard deviations across all configurations is less than 0.70\%. These results exceed those achieved using the same number of hidden layer neurons with the SKIM algorithm alone as previously reported in \cite{Cohen2016}. Combining the learnt features with the SKIM network creates a fully event-based network from end to end. The network operates on each spike, updating the features, and learning in a feed-forward manner. The SKIM network is also particularly well suited to the nature of the events produced by the adaptive threshold clustering, as they are inherently sparse spatiotemporal patterns. Where the ELM required the vectorization of the resulting spatiotemporal pattern in feature space, the SKIM network can operate on the feature events directly, and therefore has only a single input channel for each feature, allowing a smaller input layer than the ELM. This allows the largest 8000 hidden layer SKIM network to achieve the highest overall accuracy for the N-MNIST dataset. 

\subsection{Plane Dropping dataset Results}
\label{sec:plane-drop-results}

One recent example of the use of event-based sensors in a real-world application is in the field of event-based space situational awareness \cite{Cohen2017}, where the inherent wide dynamic range of event-based sensors enables detection of non-terrestrial targets both at night and during daylight. This unique advantage over traditional CCD sensors that saturate during daylight hours, together with the drastically lower data rates resulting from event-based sensing of the dark empty background of the space environment, makes event-based sensing uniquely suitable for space situational awareness. This is especially true for space-based platforms where extremely tight power and bandwidth budgets take precedence over sensor resolution, data quality, usability, and cost; areas where event-based sensors still lag too far behind traditional frame-based cameras to be commercially competitive. Yet, even in this particularly suitable environment, the unpredictability of the velocity profiles of space targets, together with the unpredictably varying temperature and lighting conditions of remote sensing environments, mean that the resultant data generated by event-based sensors are noisy and nonideal, such that often the sensor bias regime for one event polarity (and sometimes both) is entirely unsuitable for the recording environment or the observed target's velocity profile.
 
Algorithms tested and carefully tuned for ideal datasets can produce unrealistic performance expectations, and fail when tested in such challenging real-world applications. For this reason we augment our testing with a newly generated Plane Dropping dataset which provides a less controlled, more noisy dataset for classification than the N-MNIST digits dataset. It is intended to showcase the ability of the FEAST algorithm to generalize to more real-world conditions with fast, unregulated motion and unpredictable recording environments that do not match the tuned biases and controlled environments used in the generation of most event-based datasets such as N-MNIST. Additionally, although the N-MNIST dataset includes motion through saccade-like movements used to collect the dataset, this repeated tightly controlled motion profile generates repeating predictable patterns for the classifier. In the plane dropping task, the lower SNR (Signal to Noise Ratio), the varying relative orientations of the similar looking targets, and the varying velocity profiles increase the difficulty of the classification task in ways that are more similar to real world conditions.

\begin{figure*}
  \centering
  \includegraphics[width=0.80\textwidth]{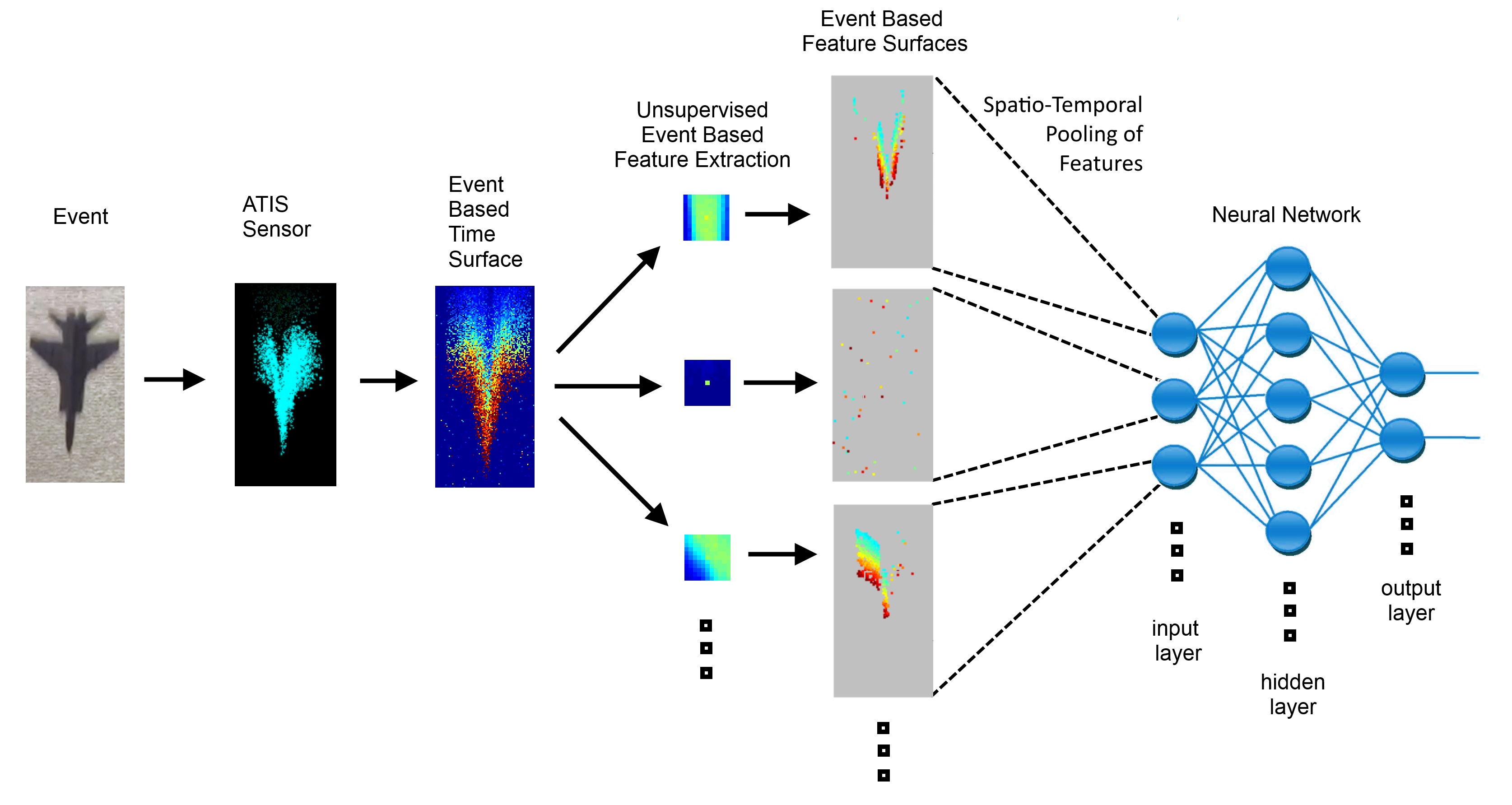}
  \caption{\textbf{FEAST Features for the Plane Dropping dataset.} The 25 spatiotemporal features produced when using the FEAST algorithm on the Plane Dropping dataset. Note that the plane dropping demonstration only uses the ON polarity due to the low SNR in the OFF events. Specific features can be seen coding for the leading edge of the airplane nosecone and for the wings. There are multiple features that code for noise events, as can be seen by several near-empty features. The colors used to represent the features are indicative of time, as these features represent complex spatiotemporal patterns. }
  \label{fig:plane-drop-system}
\end{figure*}

The algorithm used for processing the plane drop dataset was the same as that used for the N-MNIST dataset, with only three parameter modifications.

Firstly, the higher target velocities and noise levels in the Plane Dropping dataset required a shorter time constant than the N-MNIST dataset, specifically 3ms in place of 316 ms.

Secondly, whilst the same $11 \times 11$ feature size as for N-MNIST was used, the number of features selected was 25 per polarity, as networks with a higher number of features generated a large number of representations for the "noise feature" shown in Figure~\ref{fig:fspace-feature-generation}. For the Plane Dropping dataset, using $11 \times 11$ pixel features, 25 neurons consistently resulted in 2-4 variants of the noise feature which is the target range set out in the heuristic described in section~\ref{sec:Noise features and network size selection}. 

Finally, due to the non-optimized tuning of the biases of the sensor, the OFF events exhibited very low SNR and carried little information. As a result, only the ON events were used for this dataset, thereby resulting in only 25 features used in total as opposed to 200 for the N-MNIST dataset.

A diagram of the classification system used for the plane-dropping dataset is shown in Figure~\ref{fig:plane-drop-system}. The events from the camera are used to generate a time surface on which event-based feature extractors operate. The output of the feature extractors is then pooled by simply counting feature events over the entire field of view. This count is then presented to the classifier. 

As with the N-MNIST dataset, the system was trained on a subset of the airplane dataset. The training set consisted of random sets of 400 recordings, with the remaining 400 making up the test set. There is significant variance in the spatiotemporal patterns generated by the airplanes within each recording of the airplane dataset, due to significant change in velocity, pose, and the periods of partial occlusion as the planes enter and exit the field of view. This intra-recording variance significantly adds to the complexity of the dataset. To capture this variance, the feature surfaces were sampled at 3ms time intervals during each recording, resulting in approximately 50 classification operations for each recording, such that approximately 20000 unique training and 20000 unique testing samples were presented to the classifier.

\begin{figure}
  \centering
  \includegraphics[width=0.25\textwidth]{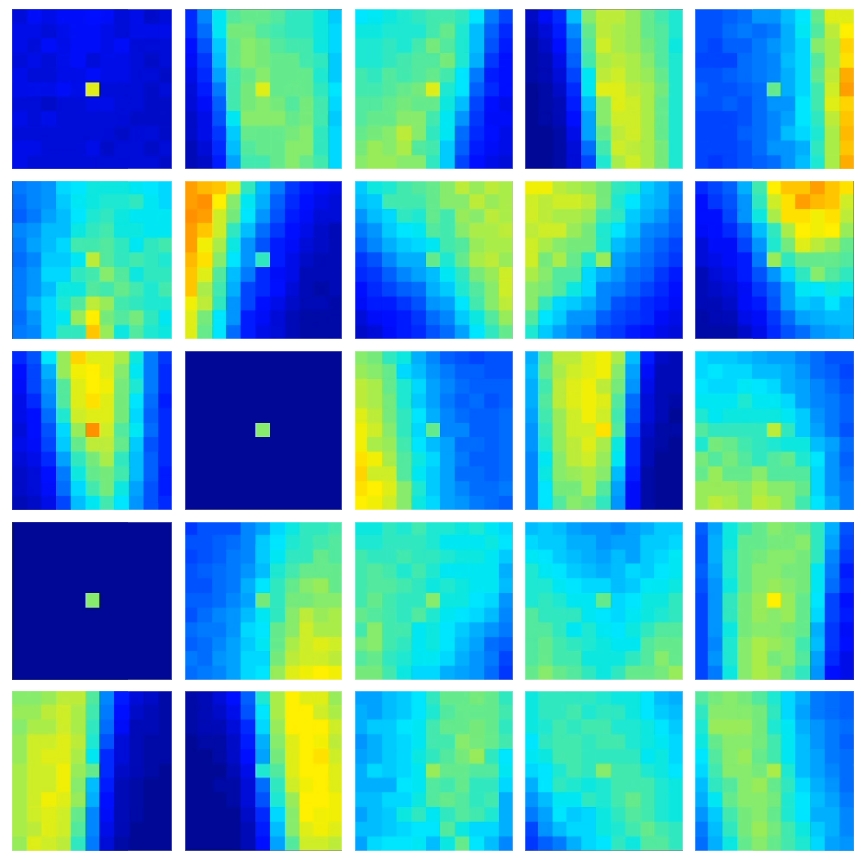}
  \caption{\textbf{FEAST Features for the Plane Dropping dataset.} The 25 spatiotemporal features produced when using the FEAST algorithm on the Plane Dropping dataset. Specific features can be seen coding for the leading edge of the airplane nosecone and for the wings. There are three features that code for noise events, as can be seen by the near-empty features.}
  \label{fig:plane-drop-features}
\end{figure}

An example of the features generated for the plane dataset is presented in Figure~\ref{fig:plane-drop-features}. These features are distinctly different from the features generated on the N-MNIST dataset as shown in Figure~\ref{fig:fspace-surface-100features}. Whereas the digit features show curves and loops, the plane features more closely resemble the slope of wings and the pointed shape of the nosecone. The network produced four variants of the noise features. Since the output of these features do not correlate with any particular class, they effectively act as naturally evolved noise detectors, leaving only clean data for the rest of the network and being essentially ignored by the classifiers. While a similar functionality can be hard coded using noise filters, the FEAST algorithm extracts multiple variants of the noise feature from event stream. These features point to subtle statistical structure in the noise which likely depends on the dynamic recording environment and the sensor. Such structured data would not be amenable to hard coding and could only be learnt and detected in an online manner.
After the convergence of the feature detector, the training data was converted to feature space through the FEAST algorithm and presented to the classifier through a supervised training regime. The training order was randomly selected. Once the training step was complete, the unseen test set was passed through the same FEAST layer and the classifier output determined. Two measures of accuracy were used. First a per frame measure calculating accuracy of each classified frame and a second per recording accuracy measure which performs a majority voting on the frames of each recording assigning the recording to the class with the highest number of winning frames.

In order to determine the effectiveness of the FEAST algorithm at mapping the input event-based data to linearly separable feature space representation, identical pooling and classification operations were performed on the raw events without any feature extraction. In addition, a linear classifier, a SKIM network with an 8000 neuron hidden layer, an ELM network of the same size, and a large 30000 hidden neuron ELM network were compared on the same feature output data in order to quantify the level of residual nonlinearity before and after the feature extraction operation. Finally, to separate the efficacy of the FEAST algorithm from the improvement gained via the event-based convolution operation, 25 random weighted features with identical weight distributions were also tested against FEAST while keeping all other aspects of the system unchanged. These results are shown in Table~\ref{tab:plane-drop-results}

\begin{table*}[]
\caption{\textbf{Summary of classification accuracies on the Plane Dropping dataset.} Four classifiers, a linear classifier, an 8000 hidden neuron SKIM network, one 8000 and one 30000 hidden neuron ELM classifier were used on the raw events, on the output of 25 random features, and on the 25 FEAST features. }
 \begin{center}
  \begin{tabular}{|l|c|c|c|c|c|c|}
  \hline
  {} & \multicolumn{3}{|c|}{Per Frame} &   
  \multicolumn{3}{|c|}{Per Drop} \\
  \cline{2-3}
  \cline{4-5}  
  \cline{6-7} 
  {Classifier} & Raw Events & Random & FEAST & Raw Events & Random & FEAST \\
  \hline
  Linear & \textbf{26.1} +/-3.1\% & \textbf{64.2} +/-4.9 \% & \textbf{83.8} +/-2.5 \% & \textbf{26.1} +/-3.1 \% & \textbf{69.6} +/-5.8 \%& \textbf{87.9} +/-2.7 \% \\ 
  SKIM 8K & N/A & N/A & N/A & N/A & \textbf{74.4} +/-5.0 \%  & \textbf{77.0} +/-3.8 \% \\
  ELM  8K & \textbf{38.3} +/-2.8\% & \textbf{67.9} +/-5.0 \% & \textbf{87.2} +/-1.9 \% & \textbf{39.1} +/-2.7\% &\textbf{75.9} +/-5.3 \% &\textbf{90.1} +/-2.2 \% \\
  ELM 30K & \textbf{40.2} +/-2.6\% & \textbf{69.2} +/-4.7 \% & \textbf{92.8} +/-1.8 \% & \textbf{41.1} +/-2.5\% &\textbf{77.8} +/-5.5 \% &\textbf{96.2} +/-2.0 \% \\
  
  \hline
  \end{tabular}
 \label{tab:plane-drop-results}
 \end{center}
\end{table*}

As the results in show, the highest per frame classification accuracy is achieved using the large ELM operating on the FEAST output, resulting in 92.81\% accuracy. More remarkable than the absolute value of the highest accuracy is the relative improvements each element of the system delivers. In the simplest baseline system, a linear classifier operates on the raw events, which, as Table~\ref{tab:plane-drop-results} shows, performs no better than chance with 26.1\% accuracy. The 8000 and 30000 hidden layer neurons of the ELM operating on the same raw events, raise this to an accuracy of 38.3 and 40.2\%, with the later likely representing the maximum achievable accuracy without use of any event-based feature extractor. Next, random features are used as feature extractors, providing a significant improvement on the ELM-on-events test, such that 64.2\% of the samples become linearly separable. The use of the large ELM on the random features only provides an additional 5\% improvement. This result provides an insight into the utility and also into the limitation of the event-based convolution operation. Despite not being effective representations of the data, the random neurons still significantly improve accuracy by aggregating local information around incoming events. Yet this aggregation is highly inefficient, with a significant amount of information lost due to the lack of specificity of the neurons to the dataset structure, such that the ELM can only extract a slight improvement on the available data despite its 30000 hidden layer neurons. In contrast, by orienting the features toward the data, the FEAST neurons alone manage to linearly separate 83.8\% of the frames and provide enough information to the ELM for it to linearize a further 9\% of the data. In this configuration of the system, when all frames of the recording are combined in a majority voting operation, a per drop accuracy of 96.2\% is achieved. Table~\ref{tab:plane-drop-confusion} details the confusion matrix for this configuration of the system for the 4-way Plane Dropping dataset.

While the SKIM classification algorithm operates on time steps that are analogous to the frames used for the linear and ELM classifier, no meaningful per frame (or per time-step) accuracy measure can be deduced from the SKIM algorithm. This is due to the nature of the algorithm which is trained to output a classification signal at the end of each recording only. In addition, operating SKIM on the raw events is extremely difficult. This is because each sensor pixel would need to be treated as an input channel, making the input layer size prohibitively large. Therefore, the SKIM network's performance was measured only on the last two measures of accuracy: per drop performance on random features and on FEAST features. On the random features the SKIM network's accuracy of 74.4\% was between the linear classifier and the tested ELMs with the equivalently sized 8000 hidden layer ELM performing slightly better than the SKIM classifier. This result is in contrast to those from the N-MNIST dataset. It is likely the result of the great variance in target velocity present in the Plane Dropping dataset compared to the near identical velocity profiles of in the N-MNIST. Here, the random kernels of the SKIM may work against the classifier by increasing the already high variance in the time scales of the observed spatiotemporal patterns caused by the varying target velocity. These differing relative performances between the classifiers highlights the utility of testing algorithms on datasets of dissimilar design.

Finally, when tested on an equal number of FEAST features that are well oriented towards the data, SKIM performs worse than a all classifiers tested. This is because the output activation of the FEAST features already provides a linearly separable mapping to the output classes but the high variance of velocity in the dataset together with the late supervisory signal in SKIM, which, on this dataset, arrives as the airplane is leaving the field of view likely impacts the algorithm's accuracy below the other per frame based methods which perform their learning at all stages of each recording providing greater invariance to target velocity.

\begin{table}[]
 \caption{Confusion matrix for mean performance of the per-frame ELM classifier on the 4-class plane dropping results.} Results averaged over 20 trials.
 \label{tab:plane-drop-confusion}
 \begin{tabular}{|l|l|c|c|c|c|r|}
           \cline{2-6}
           \multicolumn{1}{l|}{} & \multicolumn{5}{|c|}{Predicted} & \multicolumn{1}{|l}{}\\
           \hline
           &           & F117  & Mig-31 & Su-24 & Su-35 & Accuracy \\ 
           \cline{2-7}
Actual     & F117      & \textbf{24.5}\% & 0.1\%      & 0\%     & 0.5\%     & 97.7\%      \\
           & Mig-31    & 0\%     & \textbf{20.8}\%  & 0.5\%  & 3.1\%  & 85.3\%    \\
           & Su-24     & 0\%     & 0\%      & \textbf{24.79}\% & 0.07\%  & 99.1\%    \\
           & Su-35     & 0.5\%  & 2.0\%   & 0.3\%  & \textbf{21.7}\% & 88.7\%    \\
           \cline{2-6}
           & Precision & 97.9\%  & 90.8\%  & 97.0\% & 85.3\% &   \\
\hline
 \end{tabular}
 \centering
\end{table}

\subsection{Evaluating feature sets via feature activation}
The most direct measure of the utility of a feature set for any classification dataset is the recognition accuracy achieved by the classifiers. However, in many circumstances, this measure can be significantly more computationally expensive than the development of the feature set itself. Acquiring a rigorous figure of merit for any feature set can require repeated training of back-end classifiers. This long feedback loop in the evaluation of feature sets can be time consuming and can limit the range of feature extraction parameters that can be investigated. This same issue was encountered in this work, where the rigorous evaluation and comparison of feature sets through the calculation of recognition accuracy consumed significantly more time and computational resource than the development of the features themselves. However, it was found that the output of the FEAST neurons provided an easily accessed alternative signal that correlated strongly with final recognition accuracy measure. 

The adaptive selection thresholds of the FEAST neurons force the network features to capture the most commonly observed patterns, while also compensating for the frequency of the observed patterns. This means that during the learning phase the neurons are constantly being pushed toward equal activation. During inference, however, without the adaptive thresholds enforcing equal activation, the network spike rate can vary significantly across neurons with some neurons spiking more than others. This spike inequality was found to correlate strongly to the classification accuracy over the dataset, allowing rapid coarse evaluation of feature-sets and network meta parameters. The measure used for quantifying inequality in spike output was the Gini coefficient \cite{kendall1946}. This measure, commonly used to quantify wealth and income inequality \cite{sen1997} is defined as the mean absolute difference of all pairs of items in a population divided by the mean of the population to normalize the scale. The Gini coefficient $G$ is defined by ~\ref{eq:gini} where  $n$ is the number of neurons and $x_i$ and $x_j$ are the output spike counts of neurons $i$ and $j$. This measure can easily be calculated for the FEAST neurons at any point during inference. 

\begin{equation}
  \label{eq:gini}
  G = \frac{\displaystyle{\sum_{i=1}^n \sum_{j=1}^n \left| x_i - x_j \right|}}{\displaystyle{2 \sum_{i=1}^n \sum_{j=1}^n x_j}} 
\end{equation}

\begin{figure}
  \centering
  \includegraphics[width=0.50\textwidth]{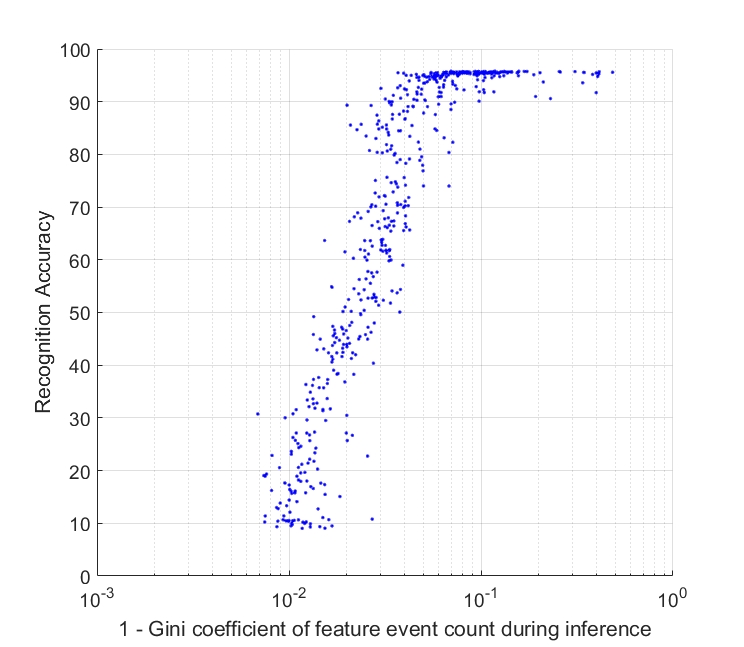}
  \caption{\textbf{Final classification accuracy and the Gini coefficient of feature event counts.} Each point on the plot represents results from an independent instantiation of a FEAST network with randomized parameters on random sections of the N-MNIST }
  \label{fig:gini}
\end{figure}

Figure~\ref{fig:gini} shows the strong, though nonlinear, relationship between the  Gini coefficient and classification accuracy for two thousand random, independently parameterized feature sets. Each data point represents results from a feature set instantiated with random threshold parameters, feature sizes, learning rates, training dataset size, and surface decay constants. The results cover almost the entire accuracy range from chance accuracy to the highest optimized accuracies achieved in this work on the N-MNIST and the Plane Dropping datasets. Yet across the entire accuracy range, the relationship between accuracy and the Gini coefficient is remarkably robust, suggesting that the Gini measure can provide a reliable rapid evaluation and comparison of feature layers, without the need for further processing and computation. Not only is the Gini coefficient useful during the algorithm design stage, where many of the interdependent parameters of the larger system need to be instantiated, but can also serve as a reinforcement signal in online learning applications, by quantifying in real-time the relevance of a feature set to any batch of observed data.

\label{sec:plane-drop-results}

\section{Discussion}

\subsection{Missed Events During Learning}
As detailed in section~\ref{sec:adaptive-threshold-clustering}, events which fall outside the threshold of all features reduce all thresholds, but do not result in an adaptation of the weights. These 'missed' events can be treated in different ways. They may be viewed as outliers with respect to the features learned by the network. This is the simplest approach in the context of hardware implementation and the one taken in this work. Another approach is to assume the missed events hold important residual information useful for classification. Being an unsupervised algorithm, the learned weights of FEAST and the output classes have no direct relationship. As such, the relative importance of unincorporated outlier events can only be determined empirically through their effect on resultant feature sets and classification performance. In all our test, the number of missed events constituted less than 5 percent of events. Experiments with a larger number of training epochs or in which missed events were re-included a second or third time into the dataset produced no observable change in the feature set or recognition accuracy. Such a result would be expected for the tests performed due to the large number of events and the significant informational redundancy in data generated by event-based sensors. Thus, while it is possible that with an extremely small and informationally sparse dataset the FEAST algorithm may not exhibit the same robustness due to missed events, this was never observed in our testing.

\subsection{Thresholds During Learning and Inference}
In our tests the initial values for the threshold were randomized. Other tests of threshold initialization included initializing the threshold at equal values at very high, or low, or zero, using uniform or Gaussian  distributions. In all such tests, no two neurons were ever detected to be in identical states, due the random initialization of the large number of weights. Because of the adaptive nature of the thresholds and the large size of the training data used, no significant difference was observed in the behavior of the signals tested across the wide range of initialization procedures and threshold adaptation parameters. In general the threshold adaptation mechanism was found to be robust to parameter selection choices, such that after a rapid initial adaptation period the thresholds of different features reached a final steady state without exception.

After training, a choice arises as to whether the selectivity information contained in the thresholds should be used during inference or simply discarded and replaced with the simple cosine distance matching rule. In this work the thresholds were disabled. Methods to incorporate the information encoded in the selection thresholds is the subject of future work.

\section{Conclusion}
\label{sec:conclusions}

The results presented in this paper demonstrate the applicability and capabilities of the FEAST algorithm for extracting useful features in an unsupervised manner. The algorithm converts the event stream into efficient feature representations which outperform random features of similar structure. The different datasets tested are shown to have significantly different feature information and noise properties. These aspects of the dataset were demonstrated in the resultant trained feature sets and used to select network size. On the N-MNIST dataset the SKIM classifier operating on FEAST features was shown to outperform all other configurations, including the ELM classifier, while on the Plane Dropping dataset the ELM on FEAST outperformed other configurations. Yet on both datasets and in all cases tested, the FEAST features outperformed raw events and random features. The adaptive selection threshold approach used in FEAST also illustrated a number of interesting properties of event-based visual classification and demonstrated the ability to perform integrated noise filtering, the generation of proxy signals for weight convergence, and ready measures for the prediction of classification performance via the Gini coefficient of the FEAST output event count.

\section*{Acknowledgment}

\ifCLASSOPTIONcaptionsoff
  \newpage
\fi

\bibliographystyle{IEEEtran}
\bibliography{Papers-FEAST}

%\begin{biography}[{\includegraphics[width=1in,height=1.25in,clip,keepaspectratio]{mshell}}]{Michael Shell}


%\vfill


\end{document}